\documentclass[twoside]{article}

\usepackage[accepted]{aistats2023}
\usepackage[round]{natbib}

\usepackage[T1]{fontenc}    
\usepackage[hidelinks]{hyperref}       
\usepackage{url}            
\usepackage{booktabs}       
\usepackage{nicefrac}       
\usepackage{microtype}      
\usepackage{xcolor}         

\usepackage{multirow}
\usepackage{amsmath}
\usepackage{amssymb}
\usepackage{amsthm}
\usepackage{amsfonts}
\usepackage{mathtools}
\usepackage{bbm}
\usepackage{caption}

\usepackage{float}

\usepackage{paralist}

\usepackage[capitalize,noabbrev]{cleveref}

\newcommand{\EE}{\mathbb{E}}

\newcommand{\RR}{\mathbb{R}}

\newcommand{\cD}{\mathcal{D}}

\newcommand{\cF}{\mathcal{F}}
\newcommand{\cB}{\mathcal{B}}

\newcommand{\by}{\mathbf{y}}

\newcommand{\balpha}{\boldsymbol{\alpha}}

\newcommand{\btheta}{\boldsymbol\theta}
\newcommand{\bdelta}{\boldsymbol{\delta}}

\newcommand{\bxi}{\boldsymbol{\xi}}
\newcommand{\bnu}{\boldsymbol{\nu}}

\theoremstyle{plain}

\newtheorem{theorem}{Theorem}

\newtheorem{assumption}{Assumption}

\newcommand*{\suppl}{Appendix}

\begin{document}

\twocolumn[
\aistatstitle{Scalable Marked Point Processes for Exchangeable and Non-Exchangeable Event Sequences}

\aistatsauthor{Aristeidis Panos \And Ioannis Kosmidis \And  Petros Dellaportas}
\aistatsaddress{University of Cambridge \And University of Warwick \\ The Alan Turing Institute \And University College London \\ Athens University of Economics and Business \\ The Alan Turing Institute} ]


\begin{abstract}
We adopt the interpretability offered by a parametric, Hawkes-process-inspired conditional probability mass function for the marks 
and 
apply variational inference techniques to derive a general and scalable inferential framework for marked point processes. 
The framework can handle both exchangeable and non-exchangeable event sequences with minimal tuning and without any pre-training. This contrasts with many parametric and non-parametric state-of-the-art methods that typically require pre-training and/or careful tuning, and can only handle exchangeable event sequences. The framework's competitive computational and predictive performance against other state-of-the-art methods are illustrated through real data experiments. Its attractiveness for large-scale applications is demonstrated through a case study involving all events occurring in an English Premier League season.

\end{abstract}





\section{INTRODUCTION}\label{sec:intro}


Point processes have been extensively used in a wide range of domains such as seismology~\citep{hawkes1971spectra,hawkes1974cluster,ogata1998space}, computational finance~\citep{bacry2015market,bacry2016estimation}, criminology~\citep{mohler2011self}, examining insurgence in Iraq~\citep{lewis2012self}, astronomy~\citep{gregory1992new}, neuroscience~\citep{cunningham2007inferring}, sports~\citep{gudmundsson2017spatio} to name a few.  

There is a voluminous literature on modelling event sequences with a vast proportion of it focusing on Hawkes process models~\citep{hawkes1971spectra} and their variants \citep{marsan2008extending,zhou2013learning,iwata2013discovering,lemonnier2014nonparametric,hansen2015lasso,xu2016learning,bacry2016first,wang2016isotonic,lee2016hawkes,eichler2017graphical,yuan2019multivariate,okawa2019deep,zhang2020variational,donnet2020nonparametric}. Most of these methods are concerned with parametric forms of the intensity functions which generalize Hawkes processes. They aim to learn, non-parametrically, the so-called triggering kernels which model the dependencies between events. There are works where the whole conditional intensity function is learned non-parametrically based on Gaussian processes~\citep{williams2006gaussian}, allowing the capture  of complex events' dynamics~\citep{liu2019nonparametric,lloyd2016latent,ding2018bayesian}. Nevertheless, interpretability is further reduced in tandem with scalability due to the computationally demanding linear algebra required for training these models. 

A more scalable solution that maintains flexibility and has produced state-of-the-art results is based on the introduction of deep learning techniques~\citep{du2016recurrent,mei2017neural,xiao2017wasserstein,li2018learning,zhang2020self,shchur2019intensity}. The majority of these methods model the intensity function of a point process through variants of recurrent neural networks (RNN). Recently,~\cite{shchur2019intensity} has proposed a method that, instead of modelling the CIF, models the conditional distribution of the inter-arrival times using a log-normal mixture density network. Their model, though, assumes independence between occurrence times and marks, which is a strong assumption for the modelling of real-world event-sequence data. Interpretability in these methods is once again limited due to the black-box nature of the neural networks. Another class of models is Graphical event models~\citep{didelez2008graphical,gunawardana2011model,bhattacharjya2018proximal} where a graphical representation of multivariate point processes is used, offering interpretability 
but suffering from scalability issues due to their squared time complexity over the number of events. A recent work~\citep{narayanan2021flexible} introduced a new family of fully-parametric marked point processes, which provides both flexibility and interpretability through a decomposition of the joint distribution over times and marks, while the resulting model had directly interpretable parameterization. Nevertheless, the inference process has been based on a Hamiltonian Monte Carlo algorithm~\citep{duane1987hybrid} in a high-dimensional space after a data wrangling procedure that eliminates parameter pre-training. Such an inference process quickly gets computationally prohibitive for large-scale event data sets.

Most of the aforementioned state-of-the-art (SOTA) methods for modelling event-sequence data are limited by the assumption of exchangeable event sequences. Despite the computational gains the exchangeability of event sequences delivers,
such an assumption may considerably reduce the flexibility of the model, failing to capture complex dynamics between event sequences. 


In this work, we inherit the interpretability offered by the conditional PMF for the marks in~\cite{narayanan2021flexible}, and we introduce an inferential framework based on variational inference (VI)~\citep{blei2017variational} that successfully deals with the aforementioned scalability limitations. We also generalize the model 
through a latent autoregressive structure over the parameters, which relaxes the exchangeability assumption. The proposed model with the autoregressive component is general, and it could find various applications where the modelling of successive event sequences is required. We demonstrate the competitive performance of our VI framework against other SOTA baselines for marked point processes on a series of real-world datasets. We also use our method to extract valuable insights over the dynamics of association football teams 
using all events in a whole association football season, which is a substantially larger data set than~\cite{narayanan2021flexible} considered. More importantly, the computational time is reduced to a few hours for data sets where~\cite{narayanan2021flexible} would require several months of computation.  

 Our main contribution is a VI-based, scalable framework for modelling marked point processes with either exchangeable or non-exchangeable event sequences. Specifically, we provide the following:
\begin{itemize}
\item  A scalable, general, VI-based inferential framework that can handle exchangeable and non-exchangeable event sequences requiring no pre-training parameter elimination and minimal tuning;
\item Competitive performance against strong SOTA baselines (e.g. based on VI and deep learning) in terms of performance, facility of training, and interoperability;
\item A large-scale case study of events from a whole association football season.
\end{itemize}


\section{MARKED TEMPORAL POINT PROCESSES FOR EXCHANGEABLE EVENT SEQUENCES}\label{sec:exchangeable_seq_theory}

\subsection{Preliminaries}\label{sec:preliminaries}
A marked temporal point process (MTPP) \citep{reinhart2018review} can be seen as an ordered sequence of event times $t_i \in [0, T)$ over an observation interval $[0, T)$, accompanied by event marks $u_i \in \mathcal{U}$, which may include information about the event types or marks  (discrete), location (continuous) or other event attributes. Our development focuses on discrete (multivariate temporal point process)  mark spaces $\mathcal{U}$, however, extension to continuous spaces (marked spatiotemporal point process) is straightforward. An MTTP is fully determined by its conditional intensity function (CIF) $\lambda(t,u | \cF_t)$ which gives the probability of observing an event in the space $[0, T) \times \mathcal{U}$ given the filtration $\mathcal{F}_t = \{ (t_i, u_i) \mid t_i < t  \}$, i.e. $\lambda(t, u \mid \mathcal{F}_t) ~||B_{du} (u) ||~ dt = \mathbb{E} \left[ N([t, t + dt] \times B_{du} (u) ) \mid \mathcal{F}_t \right]$, where $N(A)$ is the counting measure of events over the set $A \subseteq [0, T) \times \mathcal{U}$ and $|| B_{du} (u) ||$ is the Lebesgue measure over the open ball of radius $du > 0$ in $\mathcal{U}$. Given a dataset $\mathcal{D} = \{ \by_s \}_{s=1}^S$, where $\by_s = \{ (t_i^{(s)} , u_i^{(s)}) \}_{i=1}^{N_s}$ is the $s$th event sequence consisting of $N_s$ time-mark pairs $(t_i^{(s)} , u_i^{(s)})$, and under an exchangeability assumption about the event sequences, the log-likelihood of $\mathcal{D}$ is written as 
\begin{align}
  & \mathcal{L}(\mathcal{D})   = \sum_{s=1}^S \ell_s , \label{eq:data_log_lkl} 
\end{align}
where the log-likelihood of each sequence $s$ is $\ell_s = \sum_{i=1}^{N_s}  \log \lambda(t_i^{(s)} , u_i^{(s)} \mid \mathcal{F}_{t_i^{(s)}}) - \int_0^T \hspace{-0.2cm}\int_{\mathcal{U}} \lambda(t, u \mid \mathcal{F}_{t^{(s)}}) d u d t$. $\cF_{t^{(s)}} = \{ (t_i^{(s)}, u_i^{(s)}) \mid t_i^{(s)} < t^{(s)}  \}$ is the filtration of the $s$th sequence; the dependence of $\ell_s$ on the parameters and the data has suppressed for notational convenience. In what follows, we also suppress the dependence on the $s$-th sequence to reduce clutter in the notation, unless otherwise stated.  Multivariate Hawkes process \cite{hawkes1971spectra} is a well-studied MTPP, where past events contribute additively to the intensity of the current event, allowing in that way to capture mutual excitation (clustering) behaviour between events. The CIF of a multivariate Hawkes process with mark space $\mathcal{U} = \{1, \cdots, U  \}$ is given by
\begin{align}\label{eq:hawkes_cif}
    \lambda(t, u \mid \mathcal{F}_{t}) & = \mu \delta_u + \eta \sum_{j: t_j < t} \beta_{u_j,u} e^{-\beta_{u_j,u} (t - t_j)} \gamma_{u_j,u},
\end{align}
where $\mu > 0$ is a constant background intensity, $\delta_u > 0$ is the background probability for event type $u$ with $\sum_{u=1}^U \delta_u = 1$, $\gamma_{u_j,u} > 0$ is the probability of triggering an event type $u$ from the excitation of an event type $u_j$ where $\sum_{u=1}^U \gamma_{u_j,u} = 1$, $\forall u_j \in \mathcal{U}$, while $\beta_{u_j,u} > 0$ is the exponential decay rate of that excitation. The parameter $\eta \in (0, 1)$ is called the excitation factor. 

\subsection{Decoupled MTPP}\label{sec:decoupled}
Despite their widespread applicability in modelling event sequences \citep{ogata1981lewis,mohler2011self,bowsher2007modelling}, Hawkes processes perform poorly in cases where event times do not exhibit clustering behavior. An example of such a setting is football event sequences, where the inter-arrival times tend to be under-dispersed relative to a Poisson process which is the limit of a Hawkes process when $\eta$ in~\eqref{eq:hawkes_cif} approaches zero. The drawbacks of Hawkes processes in this setup are discussed in Section 3.3 of \cite{narayanan2021flexible}. These limitations are circumvented by considering the decomposition of the log-likelihood of a marked point process, 
\begin{align}
\label{eq:decomposed_llkl}
 \ell =  \sum_{i=1}^N & \left\{ \log f(u_i \mid t_i, \mathcal{F}_{t_i} ; \btheta_f) + 
    \log g(t_i \mid \mathcal{F}_{t_i} ; \btheta_g) \right\} \nonumber \\ &
    + \log \left( 1 - G(T \mid \mathcal{F}_{t_N}; \btheta_g) \right),
\end{align}
where $f(\cdot | \cdot ~; \btheta_f)$ and $g(\cdot | \cdot ~; \btheta_g)$ are the conditional probability mass function (PMF) of the event types and the conditional density for the occurrence times, respectively, parameterized by vectors $\btheta_f$ and $\btheta_g$, and $G(u \mid \cdot) = \int_{0}^u g(t \mid \cdot ) dt$. The last term in~\eqref{eq:decomposed_llkl} is the logarithm of the survival function that accounts for the fact that the unobserved occurrence time $t_{N+1}$ must be after the end of the observation interval $(0, T)$. The dependence of $\ell$ in~\eqref{eq:decomposed_llkl} on $\btheta_f$ and $\btheta_g$ has been suppressed to simplify notation and it is omitted henceforth in $f(\cdot\mid\cdot)$ and $g(\cdot\mid\cdot)$ 
unless required. All likelihood contributions in ~\eqref{eq:decomposed_llkl} share the same parameters $\btheta_f$ and $\btheta_g$, which, in turn, are assumed to have prior distributions. Therefore, the event sequences are independent conditional on  $\btheta_f$ and $\btheta_g$, and, hence, exchangeable but not necessarily marginally independent. See, for example, Section 1 in \cite{blei2003latent} for a discussion about the concept of exchangeability.

This decomposition allows defining an MTPP in terms of $f(\cdot \mid \cdot)$ and $g(\cdot \mid \cdot)$ instead of the CIF $ \lambda(t, u \mid \mathcal{F}_{t})$, providing extra flexibility to the specification of the model, and thus, added expressibility. For example, a gamma or a log-normal density could be chosen as $g(\cdot \mid \cdot)$ to capture non-clustering relations among time occurrences, like under-dispersion. 

In the current work, we adopt the same functional form of $f(u_i \mid t_i, \mathcal{F}_{t_{i-1}})$ as in \cite{narayanan2021flexible}, i.e.
\begin{equation}\label{eq:mass_function}
   f(u_i | t_i, \mathcal{F}_{t_{i}}) =  \frac{\delta_{u_i} + \eta \sum_{j: t_j < t_i} \gamma_{u_j,u_i} e^{-\beta_{u_j,u_i} (t_i - t_j) } }{1 + \eta \sum_{j: t_j < t_i} e^{-\beta_{u_j,u_i} (t_i - t_j)}},
\end{equation}
where $\eta > 0$. 
Expression~(\ref{eq:mass_function}) is obtained by converting  \eqref{eq:hawkes_cif} into a PMF by normalising across all possible values of $u$.  We define the probability vector $\bdelta = (\delta_1,  \cdots, \delta_U)^{\top}$, the stochastic matrix $\Gamma \in [0, 1]^{U \times U}$ with $\Gamma_{u,u^{\prime}} = \gamma_{u,u^{\prime}}$, the decay matrix $B \in \mathbb{R}_{+}^{U \times U}$ where $B_{u,u^{\prime}} = \beta_{u,u^{\prime}}$. Hence, the model parameters are $\btheta_f = \{ \bdelta, \Gamma, B, \eta \}$. 

The PMF of marks in \eqref{eq:mass_function} possesses a natural interpretation of its parameters. Large values of $\eta$ correspond to higher dependence of each mark on its past events since $\eta$ can be viewed as a scaling factor over the contributions of past events to the current event mark probability. The probabilities $\gamma_{u_j,u}$ can be interpreted as the conversion rates for the transition from an event type $u_j$ to an event type $u$ while the decay rate $\beta_{u, u^{\prime}}$ quantifies the exponential rate at which the excitation from a previous event with mark $u$ decays over time given the current event with mark $u^{\prime}$. The background probability $\delta_{u}$ gives the probability an event has a mark $u$ given this event is triggered exclusively by a background 
process.
In this way, we can extract useful information regarding the cross-excitations between the event types and the corresponding excitation decay rates. 

\subsection{Proposed Model for Exchangeable Event Sequences }\label{sec:model_exchangeable}
Despite the flexibility of (\ref{eq:mass_function}), the  implementation in \cite{narayanan2021flexible} is based on an MCMC algorithm that scales poorly with the number of time events after a data wrangling procedure that eliminates parameter pre-training, thus limiting its applicability only to small data sets.
We address this limitation through a VI implementation strategy that maintains the desirable properties of the model in \cite{narayanan2021flexible}, such as producing approximate posterior distributions over the parameters, while delivering vast computational speed-up over its MCMC counterpart. 
We focus on the PMF in~\eqref{eq:mass_function}, which, despite all the interpretability that brings into the model, is the computational bottleneck for the inference process.

\paragraph{Variational inference.} 
By considering a variational distribution $q_{\bxi} (\btheta_f) $, parameterized by the variational parameters $\bxi$, we aim to find these parameters that minimize the Kullback-Leibler divergence between the variational distribution $q_{\bxi} (\btheta_f) $ and the true posterior $p_{\bxi} (\btheta_f | \cD) $. This is equivalent to maximizing the evidence lower bound (ELBO) \citep{blei2017variational,zhang2018advances} defined as 
\begin{equation}\label{eq:elbo}
    \text{ELBO} (\bxi, \bnu) := \mathbb{E}_{q_{\bxi}} \left[ \log \frac{p(\cD \mid \btheta_f) p_{\bnu} (\btheta_f) }{q_{\bxi} (\btheta_f)} \right],
\end{equation}
where $p(\cD | \btheta_f) := \prod_{s=1}^S \prod_{i=1}^{N_s} f(u_i^{(s)} \mid t_i^{(s)}, \mathcal{F}_{t_i}^{(s)} ; \btheta_f)$ is the data likelihood, and $p_{\bnu} (\btheta_f)$ is the prior with hyperparameters $\bnu$, 
which are also chosen to maximize ELBO.

Regarding the variational distribution $q_{\bxi} (\btheta_f)$, we follow the mean-field approach where $q_{\bxi}$ factorizes over $\btheta_f$, i.e.
    $q_{\bxi} (\btheta_f) = q_{\bxi} (\bdelta) q_{\bxi} (\Gamma) q_{\bxi} (B, \eta)$,
where each constituent distribution is defined as
\begin{equation}
    q_{\bxi} (\bdelta) = \text{Dir} (\balpha_0), \quad 
   q_{\bxi} (\Gamma) = \prod_{u=1}^U  \text{Dir} (\balpha_u),  \label{eq:q_dirichlet_gamma}
\end{equation}
\begin{equation}
q_{\bxi} (B, \eta) = \prod_{p=1}^{U^2 + 1}  \text{Lognormal} (\mu_p, \sigma^2_p). \label{eq:q_lognormal}
\end{equation}
In the above expressions, $\balpha_p \in \RR_+^U, p = 0, 1,\cdots, U$ are the concentration parameters of each Dirichlet distribution and $\{\mu_p, \sigma^2_p \}_{p=1}^{U^2 + 1}$ are the means and variances of the log-normal distributions, and thus, $\bxi = \{ \{  \balpha_p \}_{p=0}^U, \{\mu_p, \sigma^2_p \}_{p=1}^{U^2 + 1} \}$. 

In order to maximize ELBO with respect to both $\bxi, \bnu$, we adopt a similar procedure as in \cite{salehi2019learning} where a variational EM algorithm is employed to iteratively optimize 
\eqref{eq:elbo}. For calculating ELBO, we refer to black-box variational inference (BBVI) optimization \citep{ranganath2014black,kingma2013auto} where Monte Carlo  integration is used to approximate the bound. The reparameterization  trick in \cite{kingma2013auto} allows us to obtain unbiased gradient estimates of the ELBO for the parameters of the log-normal  distributions in~\eqref{eq:q_lognormal}. However, the reparameterization  trick 
is not applicable to the concentration parameters $\balpha_i$. To overcome this issue, we make use of pathwise gradients \citep{jankowiak2018pathwise}, allowing us to 
obtain an estimate of ELBO, given by
\begin{equation}\label{eq:elbo_appx}
    \text{ELBO} (\bxi, \bnu) \approx \frac{1}{L} \sum_{l=1}^L  \log \frac{p(\cD \mid \btheta^{(l)}_f) p_{\bnu} (\btheta^{(l)}_f) }{q_{\bxi} (\btheta^{(l)}_f)},
\end{equation}
for $L$ Monte Carlo samples $\btheta^{(l)}_f$ generated from the variational distribution using the reparameterization trick/pathwise gradients. Following \cite{salehi2019learning},
 we maximize \eqref{eq:elbo_appx} with respect to the variational parameters $\bxi$ in the E-step while in the M-step we maximize  with respect to the hyperparameters $\bnu$. The updated $\bnu$ can be found in closed form when certain priors are utilized \citep{salehi2019learning}. In this work, we consider a zero-mean Gaussian prior over $\eta$ and the entries of $B$, and thus, we need $U^2 + 1$ hyperparameters to describe them. Furthermore, by imposing Dirichlet priors over $\Gamma$ and $\bdelta$, updating the concentration parameters of these priors is not required. This can be seen by writing
\begin{align*}
& \mathbb{E}_{q_{\bxi}} \left[ \log \frac{p(\cD \mid \btheta_f) p_{\bnu} (\btheta_f) }{q_{\bxi} (\btheta_f)} \right] \\ & = \mathbb{E}_{q_{\bxi}} \left[ \log p(\cD \mid \btheta_f)  \right] - \text{KL} [q_{\bxi}(\btheta_f)\mid\mid p_{\bnu}(\btheta_f)  ], 
\end{align*}
where $\text{KL}[q || p]$ is the Kullback-Leibler (KL) divergence between distribution $q$ and $p$. The above expression is maximized when the KL divergence is zero, which is true for two Dirichlet distributions when they share the same concentration parameters. Hence, these Dirichlet priors can be safely ignored from the ELBO and no hyperparameter update is required at the M-step. Details on the derivation of ELBO and the optimization procedure can be found in Section \ref{sec:appendix_derivations} of \suppl.

\paragraph{Computational speed-up.} The above model, despite its flexibility, still suffers from scalability issues due to the log-likelihood term $\log p(\cD \mid \btheta_f)$ which scales as $\mathcal{O} (S N^2)$ where $N = \max_{s=1,\cdots,S} | \by_s |$. To circumvent the problem, we assume that past events do not contribute to the evaluation of the summation term in \eqref{eq:mass_function} up to a point, and thus, they can be ignored from the computation. 
Specifically, we use the following assumptions \citep{liu2019nonparametric}:

\begin{assumption}\label{assump_1}
$\forall u =1, \cdots, U$, $\exists  B_u >0 $ such that  if $0 < B_u < t_i - t_u$, where $t_u \in \{ t_j > 0 : t_j < t_i,~  u_j = u \}$ then these events do not contribute to the sum in \eqref{eq:mass_function}.
\end{assumption}

\begin{assumption}\label{assump_2}
$\forall u =1, \cdots, U$, $\exists~C_u: \mathbb{R}_{>0} \rightarrow \mathbb{N}$ such that for any finite interval $\mathcal{A} = [t_{\rm start}, t_{\rm end}  ), | \mathcal{A} | = t_{\rm end} - t_{\rm start} $, the number of events of type $u$ in this interval $\mathcal{N}_u (\mathcal{A})$ are bounded above by  $C_u ( | \mathcal{A} |) < \infty $.
\end{assumption}

\begin{theorem}[Proof in \cite{liu2019nonparametric}]\label{theorem_1}
Under assumptions \ref{assump_1} and \ref{assump_2}, $\exists~Q \in \mathbb{N}$ such that the sum in \eqref{eq:mass_function} only requires the last $Q$ events of $\mathcal{F}_{t_{i}}$.
\end{theorem}

Theorem \ref{theorem_1} allows us to reduce the time complexity to $\mathcal{O} (S Q N)$ rendering inference feasible for large-scale datasets. This assumption can be justified by the fact that large time intervals $t_i - t_j$ lead to near to zero contributions for past events $t_j$ in~\eqref{eq:mass_function}, and hence, their absence would not affect the final log-likelihood value. $Q$ is a tunable hyperparameter that can be determined by inspecting the values of the log-likelihood on a validation dataset. Similar cut-off assumptions to speed up computations have been also considered in \cite{liu2019nonparametric} and \cite{zhang2020variational}, while a Bayesian treatment is considered in \cite{linderman2015scalable}. 

The overall complexity is affected by the number of Monte Carlo samples $M$. However, $M=1$ is usually sufficient for BBVI applications as previous studies indicate \citep{kingma2013auto,salehi2019learning}. Finally, to further reduce the computational burden, we approximate the log-likelihood~\eqref{eq:data_log_lkl} by randomly selecting batches of sequences $\cB \subseteq \{1, 2, \cdots, S \}$, and taking unbiased estimates $\sum_{s=1}^S \ell_s \approx S \sum_{s^{\prime} \in \cB} \ell_{s^{\prime}} / |\cB|.$

\paragraph{Time modelling.} Time occurrences are modelled by log-normal distributions, i.e.
   $g(t_i \mid \mathcal{F}_{t_i} ; \btheta_g) = p ( r_i \mid u_{i-1}, \tilde{\boldsymbol{\mu}}, \tilde{\boldsymbol{\sigma}})$,
where $r_i := t_i - t_{i-1}$, $t_0 = 0$,  $\Tilde{\boldsymbol{\mu}} \in \RR^U,~\Tilde{\boldsymbol{\sigma}} \in \RR^U_+$ are the means and standard deviations of the log-normal distributions, respectively, with density $p ( r_i)$. Therefore, the inter-arrival times $r_i$ are log-normally distributed where the distribution's parameters depend on the value of the previous mark $u_{i-1}$, i.e.
  $r_i \sim \text{Lognormal}(\tilde{\mu}_{u_{i-1}}, \tilde{\sigma}_{u_{i-1}}^2 )$.  
In a similar way, we could consider any continuous distribution with support on the positive reals for the inter-arrival times, such as gamma. However, we found that log-normal distributions are sufficient for the real-world datasets considered in Section \ref{sec:expers_real_world}. Note also that the distribution over inter-arrival times allows for the easy and fast prediction of the time of future events. This is in contrast to prediction for CIF-based models, which requires computationally demanding procedures, like Ogata's modified thinning algorithm \citep{ogata1981lewis}. Specifically, for a point process with CIF $\lambda(t)$, a prediction of the next time occurrence $t^*$ is computed as 
f $t^* =  \EE [\tilde{t} | \cF_t] = \int_t^{\infty} \hspace{-0.1cm} \tilde{t}~ \lambda( \tilde{t} \mid \cF_t)~ \exp \left(- \int_t^{\tilde{t}} \hspace{-0.1cm} \lambda( u \mid \cF_t) du  \right) d\tilde{t}$,
which is intractable, and Monte Carlo sampling via Ogata's algorithm is used to estimate it. In our formulation, we simply use the mode of the log-normal distribution to predict the next time occurrence. 

\subsection{Experiments on Real-World Data}\label{sec:expers_real_world}

We investigate the predictive performance of the proposed model under the exchangeability assumption in the previous section, which we term VI-Decoupled Point Process (VI-DPP), over four real-world datasets and compare the results to state-of-the-art baselines. Our code, based on PyTorch \citep{paszke2019pytorch}, is available at \url{https://github.com/aresPanos/Interpretable-Point-Processes}. 

We consider four real-world datasets with a range of characteristics, such as the number of sequences $S$, total number of events, mean sequence length etc; see Table \ref{table:stats_datasets} in \suppl. A detailed description of each of the datasets can be found in Section \ref{sec:appendix_datasets} of \suppl. Our method is trained on these four datasets and results are compared to those of three other baselines. We pick, for comparison, the VI-based Hawkes process variant in \cite{salehi2019learning} due to its state-of-the-art performance over other MLE-based methods. We use both a parametric version with an exponential triggering function (VI-EXP) and a non-parametric one with a mixture of Gaussians  triggering function (VI-SG). This method resembles ours in the way that inference is performed, however, it is limited to a parameterized form of a Hawkes process CIF and it requires careful hyperparameter tuning which is usually computationally demanding for large-scale applications. We also include in our comparisons the Self-Attentive Hawkes Process (SAHP) \citep{zhang2020self}, which is the state-of-the-art NN-based method that uses the self-attention mechanism \citep{vaswani2017attention} to address the effect of long-range/non-linear dependencies. Note that comparison with the STAN HMC implementation in \cite{narayanan2021flexible} is infeasible to the real-world datasets and the case study in Section~\ref{sec:footba_data} due to extremely high running times; see pages 80, 98 of \cite{narayanan2020football}.

\begin{table*}[ht]
\caption{Performance comparison between our proposed model (VI-DPP) and three other baselines over the four real-world datasets of Section \ref{sec:expers_real_world}. The best results across the four competing approaches are in bold.} 
\label{table:results_datasets}
\centering
\setlength{\tabcolsep}{2.2pt} 

\begin{sc}
\begin{tabular}{ll rrrrrrrr}
\toprule
\multicolumn{1}{l}{Methods}  & \multicolumn{1}{l}{Metrics} & \multicolumn{2}{c}{Mimic}   & \multicolumn{2}{c}{SOF}   & \multicolumn{2}{c}{Mooc}   & \multicolumn{2}{c}{Ret} \\
\midrule
          & LLKL    & -3.072 & (0.148) & -2.523 & (0.022)  & -1.387 & (0.024)  & -0.733 & (0.005) \\
VI-EXP   & RMSE    & 3.075 & (0.897) & 71.175 & (9.970)  & 112.805 & (6.951)  & 332.604 & (83.590) \\
          & $F_1$    & \textbf{82.28} & (5.54) & 5.52 & (0.60)  & 10.10 & (0.38)  & 41.21 & (0.52) \\
          & Time (min)    & 116.53 & (15.29) & 99.92 & (21.78)  & 842.44 & (133.20)  & 67.19 & (10.85) \\
 \midrule
          & LLKL    & -3.070 & (0.144) & -2.461 & (0.022)  & -4.260 & (0.041)  & 0.238 & (0.011) \\
VI-Gauss   & RMSE    & 5.190 & (1.817) & 53.778 & (10.098)  & 5436.984 & (486.614)  & 319.599 & (37.234) \\
          & $F_1$    & 80.77 & (4.59) & 5.64 & (0.86)  & 5.57 & (0.41)  & 40.78 & (0.82) \\
          & Time (min)    & 248.94 & (118.58) & 586.13 & (49.86)  & 1400.96 & (128.93)  & \textbf{61.57} & (17.65) \\
 \midrule
          & LLKL    & \textbf{1.534} & (0.215) & \textbf{-0.506} & (0.006)  & \textbf{1.289} & (0.136)  & 0.621 & (1.670) \\
SAHP   & RMSE    & 10.896 & (2.447) & 44.968 & (4.450)  & 28.409 & (12.342)  & 354.880 & (26.162) \\
          & $F_1$    & 63.60 & (6.38) & 8.45 & (0.31)  & 14.07 & (0.95)  & \textbf{41.31} & (0.70) \\
          & Time (min)    & 63.09 & (1.64) & 398.21 & (21.17)  & 528.34 & (26.81)  & 80.54 & (24.70) \\
 \midrule
          & LLKL    & -2.520  & (0.103) & -2.483  & (0.025)  & -0.369  & (0.014)  & \textbf{1.063}  & (0.020) \\
VI-DPP   & RMSE    & \textbf{0.969}  & (0.043) & \textbf{5.508} &  (1.235)  & \textbf{0.997}  & (0.000)  & \textbf{1.000}  & (0.000) \\
          & $F_1$    & 57.42  & (3.82) & \textbf{10.04}  & (0.24)  & \textbf{19.52}  & (1.04)  & 41.15  & (0.07) \\
          & Time (min)    & \textbf{3.45}  & (0.17) & \textbf{83.99}  & (6.03)  & \textbf{127.32}  & (1.98)  & 682.23  & (2.05) \\
\bottomrule
\end{tabular}
\end{sc}

\end{table*}

We randomly split each dataset into ten train/validation/test ($70\% / 10\% / 20\%$) sets. The train data set is used for inference, the validation dataset is used for tuning the hyperparameters, and the predictive performance is evaluated on the test dataset. Predictive performance is assessed by three different metrics: (i) mean log-likelihood (LLKL) for the ability of each method to model event sequences, (ii) root-mean-square error (RMSE) for the ability of each method to predict future time events, and (iii) $F_1$ score for the ability of each method to predict future marks. RMSE is computed as in \cite{zhang2020self}, i.e.~the square root of the sum of the squares of 
$\epsilon_i = (t_{i+1}^{*} - t_{i})/(t_{i+1} - t_{i})-1$, where $t_{i+1}^{*}$ is the predicted time of the next event. $F_1$ score is chosen to take into account possible mark imbalances. Regarding mark prediction, since our VI method gives a distribution over the parameters $\btheta_f$ and not a point estimate, the PMF in~\eqref{eq:mass_function} is computed using the mode of the log-normal distribution in \eqref{eq:q_lognormal} and the mean of the Dirichlet distributions in 
\eqref{eq:q_dirichlet_gamma}; see Table \ref{table:results_datasets} and more details in  Section \ref{sec:appendix_extra_exps} of \suppl. 

We see that VI-DPP consistently outperforms the other baselines in terms of both RMSE and $F_1$, providing evidence for the flexibility of the decoupled MTPP. SAHP scores higher values of LLKL  due to the flexibility of the NN which is based on. 
The superior performance of VI-DPP compared to the other methods in RET, MIMIC, and MOOC is due to the fact that these datasets have small inter-arrival times (or zero, to which we have to add a small positive constant as \cite{zhang2020self} do in their codebase) and/or large heterogeneity in those, resulting in either small sample means $\hat{\mu}$ of the log of the inter-arrival times and/or high variance $\hat{\sigma}^2$ of those. Hence, our mark-specific conditional log-normal models for the inter-arrival times are often trained to have modes that are very close to zero (the estimated mode of the log-normal is $\exp(\hat{\mu} - \hat{\sigma}^2)$, which results in test RMSE that is very close to 1. We also notice a significant speed-up of our method over the rest competitors for all datasets but the Retweets. This is because a small number of cut-off points $Q$ is enough to achieve good results for the first three datasets while for Retweet a larger $Q$ is required; see Section \ref{sec:appendix_extra_exps} of \suppl. It is worth mentioning that our method required minimal hyperparameter tuning in contrast to the other three baselines where meticulous hyperparameter tuning was crucial for achieving competitive results. This process leads to considerably higher computational times, especially for large-scale datasets, something that is not directly revealed in Table \ref{table:results_datasets}.
Fig. \ref{fig:retweets_plot} of \suppl  ~explores how $Q$ affects the performance of our model in terms of RMSE, $F_1$ score, and training time over the Retweets dataset. Values of $Q > 20$ do not provide any significant performance boost for both RMSE and $F_1$, while training time, as expected, increases linearly with respect to $Q$. Ignoring events in the far past, as we do for our method, cannot be applied to SAHP due to its dependence on the self-attention mechanism, and it would not improve the VI-EXP and VI-GAUSS training times because of their pre-processing steps.

\begin{table*}[ht]
\caption{Size of the association football dataset (leftmost table) and a performance comparison between our proposed model (VI-DPP) and three other baselines on the same dataset (rightmost table).} 
\label{table:size_results_dataseWootball}
{\sc 
\setlength{\tabcolsep}{4pt}
\begin{center}
\begin{tabular}{lr}
\toprule
\# of teams & 20 \\
\# of games & 380 \\
\# of sequences & 760 \\
\# of events & 524,160 \\
$U$ & 30 \\
\bottomrule
\end{tabular}
\hspace{0.9cm}
\begin{tabular}{lrrrrr}
\toprule
 & VI-EXP    & VI-Gauss    & SAHP  & THP    & VI-DPP \\
\midrule
RMSE             & 15.6430   & 14.9554     & 0.7206  & 0.6754  & \textbf{0.4855}  \\
$F_1$            & 2.43    & 2.40      & 5.64  &  16.98  & \textbf{18.73}  \\
Time (min)       & 203.77    & 145.61      &  >1200  &  >1200    & \textbf{139.91}  \\
\bottomrule
\end{tabular}
\end{center}
}
\end{table*}

\section{NON-EXCHANGEABLE EVENT  SEQUENCES}\label{sec:non_exchangeable_seq_theory}

\subsection{Relaxing Exchangeability}\label{sec:autoregressive_model}

We can relax the assumption of exchangeable event sequences
in~\eqref{sec:preliminaries} by making one or more of the parameters
in $\alpha, \bdelta, \Gamma, B$ in~\eqref{eq:mass_function} depend on
the event sequences or on a set of event sequences. Then, we can
assume autoregressive latent processes for those parameters. Here, we
focus on placing autoregressive latent processes on the conversion
rates $\Gamma$ because of the facility to incorporate process- or
event-specific covariate information whose effect is directly
interpretable in terms of log-odds of triggering an event type from
the excitation of another event type. 

The event-sequence-specific conversion rates can be linked
to covariate vectors $(x_{t1}, \ldots, x_{tp})^{\top}$
observed at time $t$ by letting
\[
\log \frac{\gamma_{u,u^{\prime}}(t^{(s)})}{\gamma_{u, U}(t^{(s)})} =
\phi_{u,u^{\prime}} + \sum_{j = 1}^p\omega_{ju^{\prime}}^{(s)}
x_{t^{(s)}j},
\]
$\forall~u^{\prime} \in \{1, \cdots, U-1\}, u \in \{1, \cdots, U\}$. We then assume that the covariate effects
follow independent autoregressive processes with
$\omega_{ju^{\prime}}^{(1)} \sim \mathcal{N}( \mu_{ju^{\prime}} ,~
\sigma_{ju^{\prime}}^2 / (1 - \rho_{ju^{\prime}}^2))$ and
$\omega_{ju^{\prime}}^{(s)} \mid \omega_{ju^{\prime}}^{(s - 1)} \sim
\mathcal{N}( \mu_{ju^{\prime}} + \rho_{ju^{\prime}}
\omega_{ju^{\prime}}^{(s-1)},~ \sigma_{ju^{\prime}}^2 )$
$(s = 2, \ldots, S)$. The log-likelihood is given by
\begin{align}      
\log \int_{\boldsymbol{\omega}} \prod_{s=1}^{S} ~ \mathcal{L}^{(s)}( \cF^{(s)} , \boldsymbol{\omega}^{(s)} ) ~d P(\boldsymbol{\omega}) ,
\label{eq:likl_autoregressive}
\end{align}
which, in contrast to~\eqref{eq:data_log_lkl}, is not tractable because we need to integrate out the latent parameters.
$\mathcal{L}^{(s)}( \cF^{(s)} , \boldsymbol{\omega}^{(s)} )$ is the
event-type likelihood from all events in the  $s$-th event sequence
defined by the mass function in~\eqref{eq:mass_function}, and
$\cF^{(s)}$ is all the information available up to and including the $s$-th event sequence. The prior probability measure $P(\boldsymbol{\omega})$ is a
$p \times (U - 1) \times S$-dimensional Gaussian measure with mean
vector defined by $\mu_{ju^{\prime}}$ and covariance matrix fully described by $\rho_{ju^{\prime}}$ and
$\sigma_{ju^{\prime}}$; more details about 
autoregressive processes can be found in Section 2 of
\cite{rue2005gaussian}. The logarithm is outside the integral in
\eqref{eq:likl_autoregressive}, making its computation numerically
unstable. Hence, we resort to Jensen's inequality to obtain a lower
bound on \eqref{eq:likl_autoregressive},
\begin{align}
  \sum_{s=1}^{S}~ \int_{\boldsymbol{\omega}^{(s)}}  ~ \log \left( \mathcal{L}^{(s)}( \cF^{(s)} \mid \boldsymbol{\omega}^{(s)} ) \right) ~d P(\boldsymbol{\omega}^{(s)}) , \label{eq:lb_likl_autoregressive}
\end{align}
which holds due to logarithm properties and conditional independence
of $\omega_{ju^{\prime}}^{(s)}$.  By treating
$\{ \mu_{ju^{\prime}} ,~ \sigma_{ju^{\prime}},~ \rho_{ju^{\prime}} \}$ 
as hyperparameters, we can naturally apply the VI framework of Section
\ref{sec:model_exchangeable} by plugging
\eqref{eq:lb_likl_autoregressive} into \eqref{eq:elbo_appx} and
maximize the new variational lower bound with respect to
hyperparameters, the variational parameters, and the whole vector
$\boldsymbol{\omega}$, and thus, deriving an approximation over the
posterior mode of $\boldsymbol{\omega}$ given all the association
football data.

Notice that the ELBO is regularized by the log-Gaussian density over
the latent $\boldsymbol{\omega}$. The optimization procedure is now
identical to the one followed in Section \ref{sec:model_exchangeable}
while the assigned variational densities and priors for
$\btheta_f = \{ \bdelta, \Gamma, B, \eta \}$ hold the same as
well. The $p \times (U - 1)$-dimensional Gaussian integrals in
\eqref{eq:lb_likl_autoregressive} are approximated by Monte Carlo
integration since the convenient structure of the inverse covariance
matrix of an autoregressive process allows sampling with linear time
complexity \citep{rue2005gaussian}.

As is done in the application of the next section, the above applies unaltered if the non-exchangeable event sequences are replaced by non-exchangeable sets of event sequences.

\subsection{Association Football Data}\label{sec:footba_data}

Over the last decade, the importance of analysing association football matches, in tandem with the availability of spatiotemporal data from these matches, have sparked the development of many research works focusing on the statistical modelling of teams' and/or players' performance. Recent works on the analysis of spatiotemporal data from team sports, such as football, have been developed with a primary focus on each player's performance or the pattern extraction in the team plays. These works use for their analysis, as we do in this section, event data streams from various team sports where each event can be fully described by the occurrence time, its location, its type (goal, pass, etc.), and the involved players, and team information. For instance, \cite{passos2011networks,grund2012network,duch2010quantifying,clemente2015general} focus on modelling player interaction through network analysis. Other works
aim to identify patterns from pass sequences \citep{wang2015discerning,van2016strategy} or other patterns that lead to a goal event \citep{decroos2017predicting}. 

Another stream of literature relies on the extraction of game states from event sequences in order to numerically assess in-game player actions
\citep{routley2015markov,decroos2019actions} or to predict goal probabilities given the current game state \citep{robberechts2019will}. \cite{gudmundsson2017spatio} provides a detailed survey of the use of spatiotemporal analysis in team sports.  A more recent work~\citep{narayanan2021flexible} followed a different direction where they studied the dynamics of association football matches by modelling all event sequences within a game through a marked point process. The framework in~\cite{narayanan2021flexible} relies on MCMC in a high-dimensional space after a data wrangling procedure that eliminates parameter pre-training, and can only handle exchangeable event sequences like most of the state-of-the-art (SOTA) methods for modelling event-sequence data. This poses severe computational and conceptual limitations in the realistic modelling of large-scale event-sequence data.

\begin{figure*}[ht]
\centering
\includegraphics[width = \textwidth]{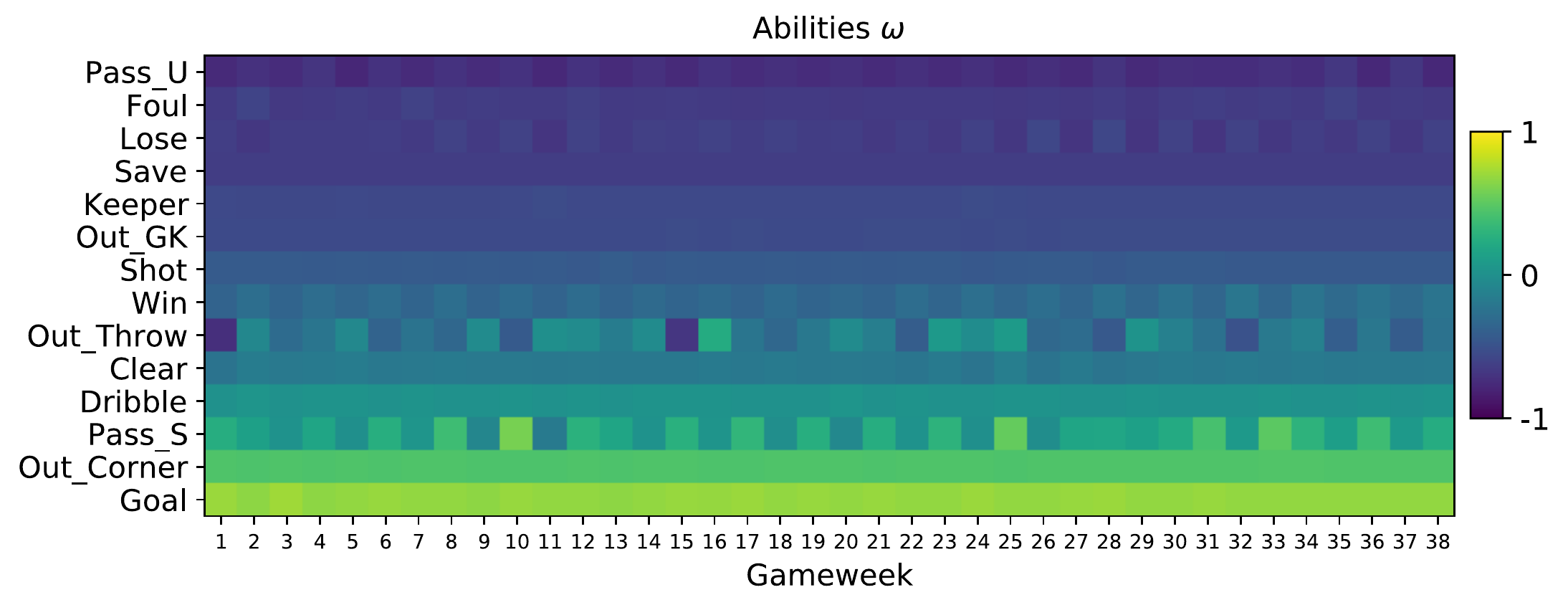} 
\caption{Values of Spearman's rank correlation coefficient between $\omega_{\cdot, u^{\prime}}^{(w)} \in \mathbb{R}^{20}, ~w=1\cdots, 38$, for each event type $u^{\prime}$ and the accumulated points awarded to each team at the end of the season, throughout the 38-week season. }
\label{fig:omegas_spearman}
\end{figure*}

\begin{figure*}[ht]
\centering
\includegraphics[width = \textwidth]{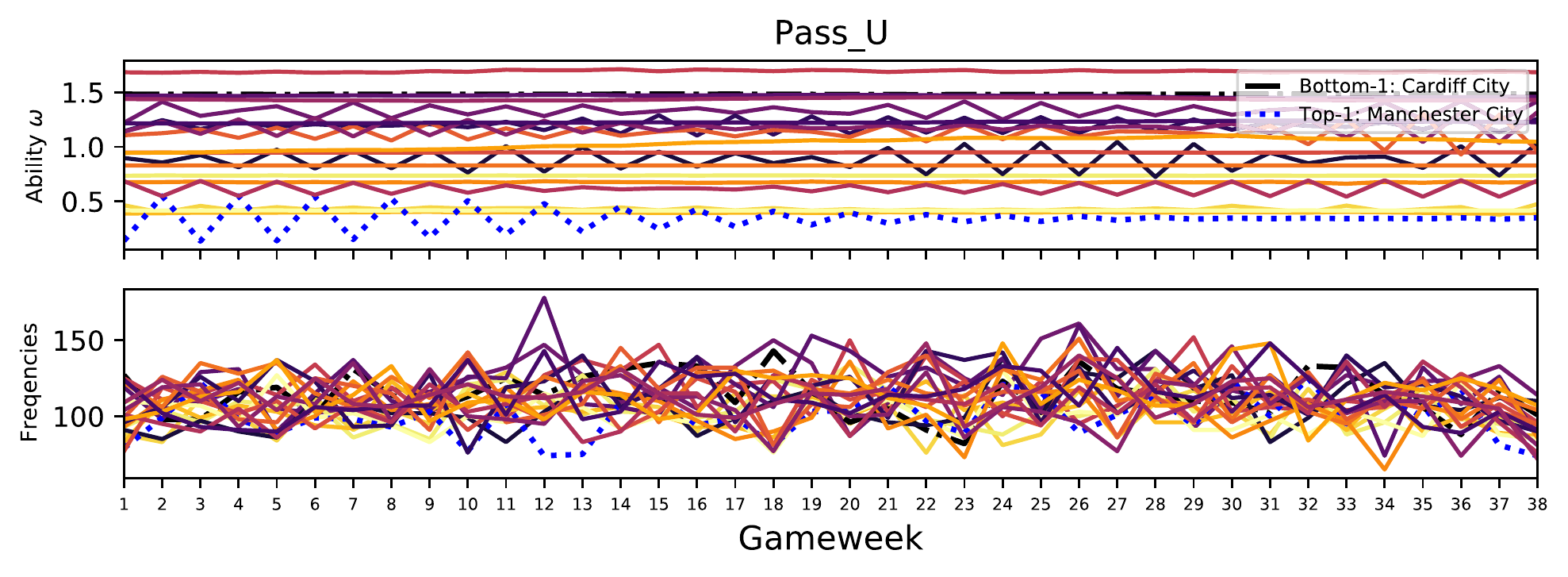} 
\caption{Top panel shows the evolution of unsuccessful passes (Pass\_U) ability $\omega_{c, 3}^{(t)}$ for each team across the season while the bottom one illustrates the same information using the raw frequencies of Pass\_U from the association football dataset. The blue and black dashed lines correspond to Manchester City and Cardiff,  the winner and bottom-ranked teams of the 2013/2014 Premier league respectively. The colors for the rest of the teams have been assigned according to their final rank, with brighter colors associated with teams with higher ranks at the end of the championship. }
\label{fig:omegas_pass_u}
\end{figure*}

The data\footnote{The football dataset used in Section~\ref{sec:footba_data} is proprietary and we do not have the license to make it publicly available.} used in this study consists of all touch-ball events recorded in all English Premier League (EPL) games throughout the season 2013/2014. Each sequence includes the touch events from one of the two halves of each game. Hence, we have 760 sequences for 380 games between 20 teams for the season and more than half a million touch-ball events. The dataset consists of triplets $(t, u, z)$, where $t$ is the time when the touch-ball event occurred, $u$ is the event type with $u \in \{1, \ldots, 30\}$  and $z \in \{ 1, 2, 3  \}$ denotes the spatial location, or zone, in the football field that the event took place. 
There are 15 distinct types labeled by which team (home or away) triggered the event; see Table \ref{table:football_description} in \suppl{} and \cite{narayanan2021flexible} for a detailed description of the data and pre-processing steps. 

In association football, the process ends immediately after the last event in each half of the game. Hence, the last term in~\eqref{eq:decomposed_llkl} is not part of the likelihood; see Section~4.2 of  \cite{lindqvist2006statistical}.

We consider a generalization of~\eqref{eq:mass_function} from \cite{narayanan2021flexible} where

\begin{align}
\label{eq:mass_function_football}
        f(u_i | t_i, z_i, \mathcal{F}_{t_i} )  = 
 \frac{\delta_{u_i}^{z_i} + \alpha \sum_{j: t_j < t_i} \gamma_{u_j,u_i}^{z_i} e^{-\beta_{u_j,u_i}^{z_i} (t_i - t_j) }}{1 + \alpha \sum_{j: t_j < t_i} e^{-\beta_{u_j,u_i}^{z_i} (t_i - t_j) }},
\end{align}
that accounts for the zone information. 

The parameters
$\bdelta, \Gamma, B$ are now location specific, i.e.
$\bdelta^z_u = \delta_u^z, \Gamma^z_{u, u^{\prime}} =
\gamma_{u,u^{\prime}}^z$, and
$B^z_{u, u^{\prime}} = \beta_{u,u^{\prime}}^z$. Team information is
incorporated in the model via the baseline-category logit
representation
$\log \{ \gamma_{u,u^{\prime}}^z(t) / \gamma_{u, U}^z(t) \} =
\phi_{u,u^{\prime}}^z + \omega_{c(t) u^{\prime}}^{(w(t))}$
$\forall~u^{\prime} = 1, \cdots, U-1,~c=1, \cdots, 20,$ where $c(t)$ is the index of the team associated with the touch-ball event at time $t$, $w(t) \in \{1, \cdots, W\}$ is the game week of the event at time $t$, where $W$ is the number of game weeks in the season, $\phi_{u,u^{\prime}}^z$ is a location-specific base parameter, and
$\omega_{c u^{\prime}}$ reflects the ability of the team $c$ to
complete a conversion to an event of type $u^{\prime}$.
For the autoregressive latent process, we assume that
$\omega_{c, u^{\prime}}^{(1)} \sim \mathcal{N}( \mu_{c, u^{\prime}} ,~
\sigma_{c, b(u^{\prime})}^2 / (1 - \rho_{c, b(u^{\prime})}^2) )$, and
$\omega_{c, u^{\prime}}^{(w)} \mid \omega_{c, u^{\prime}}^{(w-1)} \sim
\mathcal{N}( \mu_{c, u^{\prime}} + \rho_{c, b(u^{\prime})} \omega_{c,
  u^{\prime}}^{(w-1)},~ \sigma_{c, b(u^{\prime})}^2 )$,
and $b(u^{\prime}) = \max(1, u^{\prime} \bmod 16)$, i.e. we
choose different means $\mu_{c, u^{\prime}}$ for both home and away
event types while the 15 distinct (home and away) event types share
common $\rho_{c, b}$ and $\sigma_{c, b}$. This choice is justified
from the fact that each team, typically, plays one game home and one away in the next game week.

\subsection{Experimental Results}\label{sec:football_results}
We trained the new extended model using the full association football dataset (Table \ref{table:size_results_dataseWootball}) with $33,932$ parameters being learned. We obtained the optimized $\boldsymbol{\omega}$, which has a natural interpretation, and we investigated possible patterns of its values throughout the whole season. We probed whether the various abilities are related to the final ranking of the teams at the end of the season by computing Spearman's rank correlation coefficient between each one of the abilities and the final accumulated points earned by each team. Figure \ref{fig:omegas_spearman} illustrates that event types such as Goal and Pass\_U seem to be strongly correlated with the final league table of the championship from the very first game weeks. 

Figure \ref{fig:omegas_pass_u} illustrates that abilities are more informative than the raw frequencies of the event type Pass\_U for each game week. While noisy frequencies are hard to interpret across the season,  the evolution of the Pass\_U ability clearly indicates that teams with low ability of unsuccessful passes are found in the top ranks of the championship for each game week. Manchester City (MC) seems to have a more variable ability in the first game weeks of the season, possibly due to the alternate home/away games, but its ability stabilizes after the 20-th game week. Notice the strong correlation of Pass\_U with the final ranking where teams with higher ranks (lighter colors) are close to the championship winner MC. Similar patterns for other event types are illustrated in Figure \ref{fig:omegas_lose} of \suppl.

We have extended the code of the three competing methods of Section \ref{sec:expers_real_world} so that the $\log$-survival term is not included in their log-likelihood computations and we compare their performance with our proposed method on the association football dataset. 
We have also added another baseline (THP) based on transformer architecture \citealp{zuo2020transformer} in our comparisons on the football dataset. We use the first 37 game weeks for training and the last game week for model evaluation, see Table \ref{table:size_results_dataseWootball}; VI-DPP outperforms all the baselines across all three metrics.  Out-of-sample log-likelihood values are not presented since the computation of \eqref{eq:likl_autoregressive} is intractable for our model and thus, no direct comparison to the other baselines would be possible. 
We attribute the superior performance of the non-exchangeable version of VI-DPP over its competitors to the fact that the competing methods either assume that the event sequences are independent, e.g. see Section 5 in \cite{zhang2020self} for SAHP, or that there is a single event sequence, e.g. see \cite{salehi2019learning} for VI-EXP and VI-Gauss. Assuming exchangeability or independence effectively ignores the ordering of the event sequences, which is a restrictive assumption in settings like the analysis of football games, where it is reasonable to expect that team abilities vary during the season.

Our model can also produce event genealogies, which allow the probabilistic identification of the events in the past that are most likely to trigger a present event of interest, such as a goal. Section \ref{sec:branching} of \suppl{} provides a detailed discussion about the computation and visualization of event genealogies post-training. Furthermore,  interpretable results for association football games are presented based on the corresponding computed event genealogies. These results are supported by links to the actual footage of these games in the data, which confirms the insights extracted by our method.



\section{DISCUSSION}\label{sec:discussion}
We have proposed a novel inferential framework for a flexible family of interpretable MTPPs based on VI enjoying scalability benefits, under the assumption of exchangeable event sequences. We have also presented an extension of this model that accounts for successive event sequences, and thus, generalizing the work of \cite{narayanan2021flexible}, with the goal of modelling association football in-game events. Nevertheless, other applications of the model could be considered where modelling of successive event sequences is required. A case study based on a large volume of association football events data has been demonstrated where the scalability and interpretability of our method has lead to valuable insights of event and team dynamics for the whole season. Experiments on real-world datasets illustrate that our framework has competitive performance over recent baselines, some of which involve neural network specifications. The usefulness of the framework becomes more apparent due to its  minimal hyperparameter tuning, which is in contrast to the other three baselines where meticulous hyperparameter tuning is crucial for achieving competitive results. 

The main limitation of the VI-DPP model, as defined here, is that it cannot capture inhibition behavior \citep{costa2020renewal,chen2017multivariate,bonnet2021maximum}, i.e. having the occurrence of an event decrease the likelihood of another event to occur. This mainly concerns the mark space, where due to the additive nature of \eqref{eq:mass_function}, the appearance of a mark increases the likelihood of another one triggering. Nevertheless, our experimental evaluation shows that our model is flexible enough to capture the dynamics of various complex real-world data. 


\subsubsection*{Acknowledgements}

We thank the reviewers for their insightful comments that enabled us to improve key aspects of our work. This work was supported by the Bill \& Melinda Gates Foundation [INV-001309] through the "Trustworthy digital infrastructure for identity systems" project of The Alan Turing Institute.


\bibliography{refs}
\bibliographystyle{plainnat}

\appendix
\onecolumn

\section{ON THE DERIVATION OF VI-DPP}\label{sec:appendix_derivations}
The ELBO defined in \eqref{eq:elbo} can be also seen as a lower bound on the log-marginal likelihood $$\log p (\cD; \bnu):= \log \int p(\cD | \btheta_f) p_{\bnu} (\btheta_f)~d\btheta_f, $$
where a direct application of Jensen's inequality gives
$$ \log p (\cD; \bnu) \geq \text{ELBO} (\bxi, \bnu).$$
Maximizing ELBO with respect to $\bxi$ gives a tighter bound on the log-marginal likelihood. Hence, a variational EM algorithm, similar as the one described in \cite{salehi2019learning}, can be used to efficiently optimize the variational parameters $\bxi$ and the hyperparameters $ \bnu$. At the E-step, the ELBO is maximized with respect to $\bxi$, giving in that way a better approximation to the log-marginal likelihood, and then, at the M-step, the updated ELBO is maximized with respect to $\bnu$. Nevertheless, we found empirically that the optimization of ELBO converges faster to a local maximum when the M-step is ignored and no prior information is incorporated into our VI framework. We postulate this behavior stems from a large amount of data available in our experiments which provides enough information to train our model efficiently without any prior information needed. Hence, the only parameters we need to optimize are the variational parameters $\bxi$ using the following objective function
\begin{equation}\label{eq:real_objective}
\mathbb{E}_{q_{\bxi}} \left[ \log p(\cD | \btheta_f) \right] \approx  \frac{1}{L} \sum_{l=1}^L  \log p(\cD | \btheta^{(l)}_f).
\end{equation}
This objective function is the same as the ELBO in \eqref{eq:elbo_appx} without the regularization term $\text{KL} [q_{\bxi}(\btheta_f)~||~p_{\bnu}(\btheta_f)  ]$, which can be obtained when the chosen prior $p_{\bnu}(\btheta_f)$ is identical to the variational distribution $q_{\bxi}(\btheta_f)$. Since the choice of prior has negligible importance in the presence of large amount of data and empirical evidence showed that faster convergence is attained by ignoring the KL-term, we opted to optimize only the variational parameters using \eqref{eq:real_objective}.

Unlike typical Bayesian inference, in variational inference, it is customary to optimize over the prior parameters, instead of fixing them before seeing the data. Specifically, this is achieved through optimizing the ELBO in \eqref{eq:elbo} over the prior parameters and variational parameters jointly; for example, see  Eq.~(12) in Salehi et al. (2019), where their work concerns a similar context. In our formulation, the variational and prior distributions for $\boldsymbol{\delta}$ are both Dirichlet with different parameters. Hence, during the M-step of the ELBO maximization procedure, for any fixed $\boldsymbol{\xi}$ the KL divergence of $q_{\boldsymbol{\xi}}(\boldsymbol{\delta})$ from $p_{\boldsymbol{\nu}}(\boldsymbol{\delta})$ achieves its global minimum of zero when the sub-vector of $\boldsymbol{\nu}$ corresponding to $\boldsymbol{\delta}$ is exactly equal to the sub-vector of $\boldsymbol{\xi}$ corresponding to $\boldsymbol{\delta}$. The same holds for $\Gamma$.

\section{EXTRA EXPERIMENTAL DETAILS}\label{sec:appendix_extra_exps}
The exact experimental setups for each of the methods used in Section \ref{sec:expers_real_world} are discussed here. 

All methods were trained for 2000 epochs using batches of size 32 and setting as default optimizer Adam \cite{kingma2014adam}. We also used the log-likelihood of the validation dataset for early stopping a method that does not improve its log-likelihood for a hundred consecutive epochs. This was not necessary for the three VI-based methods since no sign of overfitting was observed. However, this was crucial for the NN-based SAHP where overfitting was common in all four datasets.

Regarding hyperparameter tuning, each method has its own set of hyperparameters that requires thorough tuning. The choice of each set of hyperparameters was based on the configuration that maximized the log-likelihood on the validation dataset by grid-searching over the parameter space. More accurately, for each method we have:

\paragraph{VI-EXP.} For each dataset, we tried decay $\in \{0.1, 0.5, 1, 2, 4, 8, 16, 32 \}$. We chose $\text{decay} = 2$ for MIMIC dataset, decay=0.1 for SOF, decay=16 for MOOC, and decay=0.1 for Retweets.

\paragraph{VI-GAUSS.} Here we had to tune two hyperparameters, the number of Gaussian basis $M$ and the cut-off time $T_c$. The center of the $m$-th Gaussian kernel is $t_m = T_c \cdot (m-1) / M$ and its scale is given by $s = T_c / (\pi M)$; see \cite{salehi2019learning} for more details. For each dataset, we tried $M \in \{1, 2, 4, 10, 15, 20, 30 \}$ and $T_c \in \{M/4, M/2, M, 2M, 3M, 4M, 5M, 8M, 16M \}$. We chose $M=2, T_c=4$ for MIMIC dataset, $M=20, T_c=60$ for SOF, $M=4, T_c=64$ for MOOC, and $M=30, T_c=150$ for Retweets.

\paragraph{SAHP.} Hyperparameter tuning was the most challenging one for this model since it comes with a large number of hyperparameters due to its neural net dependence. For example, we have the number of hidden units, number of layers, number of attention heads and dropout ratio for the neural net's weights. We also found it important to set a warm-up schedule to increase and then reduce the learning rate throughout optimization as suggested in \cite{zhang2020self}. We found that four hidden layers, four attention heads, dropout=0.1, and initial learning rate equals to $3 \times 10^{-5}$  worked well across all datasets and thus we kept that values fixed. We chose 32 hidden units for the MIMIC dataset, 128 for both SOF and MOOC, and 64 for Retweets.

\paragraph{VI-DPP.} For our method the only parameter needed tuning  was the number of cut-off points $Q$. We chose $Q=1$ for all datasets except Retweets where $Q=15$ was used. Other parameters such as momentum term and the number of MC samples $L$ were set as in \cite{salehi2019learning}, i.e. 0.5 for momentum term and $L=1$.

Regarding the learning rate for VI-EXP/GAUSS, it was set $0.05$ and kept fixed over all datasets. Similarly, for VI-DPP a common learning rate was used with a value $0.03$

\begin{figure*}
\centering
\includegraphics[width=14.1cm,height=4.8cm]{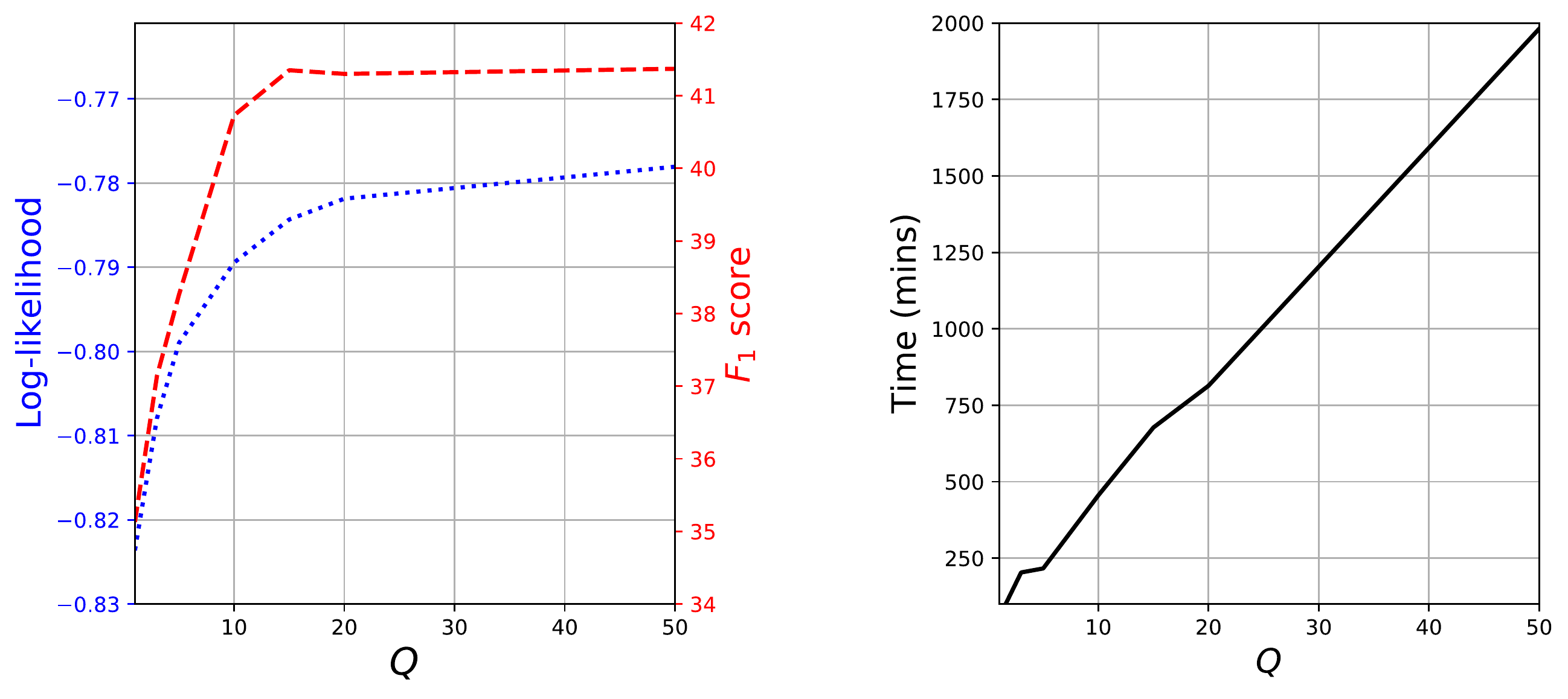} 
\caption{The values of log-likelihood (left panel, blue dotted line), $F_1$ score (left panel, red dashed line), and training time (right panel, black solid line) as a function of the number of  cut-off points $Q$ for our model VI-DPP trained over the Retweet dataset. The log-likelihood computation here takes into account only the marks.}
\label{fig:retweets_plot}
\end{figure*}

\begin{table*}[ht]
\caption{Mark type description for the association football data. The value of p can be either ``H'' or ``A'' depending on whether an event is triggered by the home or away team, respectively. The events with the ``H'' prefix have $u \in \{1, \cdots, 15 \}$ while ``A'' events have $u \in \{16, \cdots, 30 \}$. If $u$ is a ``H'' event then $u+15$ is the corresponding ``A'' event, and thus, we have $U=30$ distinct event types.}
\label{table:football_description}
\centering
\begin{tabular}{| c | c | c  | } 
\hline
$u$ & \textbf{Label Name} & \textbf{Description}\\ 
\hline
1 or 16 & p\_Win &  A player of the p team regains possession of the ball from the \\
 & & opponent. \\
\hline
2 or 17 & p\_Dribble &  A player of the p team takes the ball forward with repeated slight \\
& & touches. \\
\hline
3 or 18 & p\_Pass\_S &  A player of the p team gains possession of the ball from a pass\\
 & & coming by one of his teammates. \\
\hline
4 or 19 & p\_Pass\_U &  A player of the p team failed to pass successfully the ball to one of \\
 & & his teammates. \\
\hline
5 or 20 & p\_Shot &  A player of the p team shots the ball at the opponent's goal. Attempts \\
 & & where the ball misses the target are also included. \\
\hline
6 or 21 & p\_Keeper &  The goalkeeper of the p team takes possession of the ball into their   \\
 & &  hands by picking it up or claiming a cross. \\
\hline
7 or 22 & p\_Save &  The goalkeeper of the p team prevents a shot from crossing the \\
 & & goal line. \\
\hline
8 or 23 & p\_Clear &  A player of the p team moves the ball away from his goal area to \\
 & & safety. \\
\hline
9 or 24 & p\_Lose &  A player of the p team loses possession of the ball. \\
\hline
10 or 25 & p\_Goal &  A player of the p team scores a goal. \\
\hline
11 or 26 & p\_Foul &  A player of the p team executes a free-kick due to a previously \\
 & & occurred foul. \\
\hline
12 or 27 & p\_Out\_Throw &  A player of the p team sends the ball out-of-play. \\
\hline
13 or 28 & p\_Out\_GK &  The goal keeper of the p team sends the ball out of play. \\
\hline
14 or 29 & p\_Out\_Corner &  A player of the p team  sends the ball out of play over the p team's \\
 & & goal line. \\
\hline
15 or 30 & p\_Pass\_O &  A pass from a player of the p team to one of his teammates who is   \\
& & judged guilty of the offside offence. \\
\hline
\end{tabular}
\end{table*}

\section{DATASETS}\label{sec:appendix_datasets} We provide a short description on the four real-world datasets used in Section 3.4 of the main paper while  quantitative characteristics of these datasets are given in Table \eqref{table:stats_datasets}.

\paragraph{MIMIC-II (MIMIC).} The Multiparameter Intelligent Monitoring in Intensive Care (MIMIC-II) is a medical dataset consisting of clinical visit records of intensive care unit  patients for seven years. There are records of 650 patients/sequences where each one contains the time of the visit and the diagnosis result of this visit. There are $U=75$ unique diagnosis results. The goal is to predict the time and the diagnosis result of a patient. 

\paragraph{Stack Overflow (SOF).} The data comes from the well-known question-answering website Stack Overflow\footnote{\url{https://archive.org/details/stackexchange}} where users are encouraged to answer questions so they can earn badges. There are $U=22$ different types of badges. The data have been obtained from 01/01/2012 to 01/01/2014. Each sequence corresponds to a user and each event gives the time and the type of budge a user has been awarded.

\paragraph{MOOC.} This dataset consists of the interactions of students on a massive open online course (MOOC) on XuetangX\footnote{\url{http://moocdata.cn/challenges/kdd-cup-2015}}, one of the largest MOOC platforms in China. The interactions are $U=98$ in total, and some examples are video viewing, answer submission etc. A sequence contains the interactions with their corresponding occurrences of a given user.

\paragraph{Retweets (RET).} The Retweets dataset includes retweets sequences, each commencing with an original tweet. Each retweet is described by the time it occurs and the type of this retweet; we have $U=3$ retweet types: small, medium and large ones, depending on the popularity (number of followers) of the retweeter. The aim is to predict when the next retweet will be and how popular the next retweeter will be.

\begin{table*}[ht]
\caption{Characteristics of the real-world datasets. $U$, $S$, $\#$, are the numbers of marks, sequences, and events (in thousands), respectively. TR, VA, and TE are training, validation, and test, respectively.} 
\label{table:stats_datasets}
\begingroup
\setlength{\tabcolsep}{3.3pt} 
\renewcommand{\arraystretch}{0.8}
\begin{small}
\begin{center}
{\sc
\begin{tabular}{l rrrrrrrrrr}
\toprule
Dataset & \multirow{2}{*}{$U$} & \multicolumn{3}{c}{Sequence lengths} & \multicolumn{3}{c}{$S$} & \multicolumn{3}{c}{$\#$} \\
\cmidrule(lr){3-5}  \cmidrule(lr){6-8}  \cmidrule(lr){9-11}
 &  & \multicolumn{1}{c}{Min} & \multicolumn{1}{c}{Max} & \multicolumn{1}{c}{Mean} & \multicolumn{1}{c}{TR} & \multicolumn{1}{c}{VA} & \multicolumn{1}{c}{TE} & \multicolumn{1}{c}{TR} & \multicolumn{1}{c}{VA} & \multicolumn{1}{c}{TE} \\
\midrule
Mimic \cite{johnson2016mimic,du2016recurrent} & 75 & 2 & 33 & 4 & 454 & 66 & 130 & 1.6 & 0.25 & 0.5 \\
SOF \cite{du2016recurrent} & 22 & 41 & 736 & 72 & 4643 & 663 & 1327 & 336 & 47 & 95 \\
Mooc \cite{kumar2019predicting} & 97 & 4 & 493 & 56 & 4932 & 705 & 1410 & 279 & 37 & 79 \\
Ret \cite{zhao2015seismic} & 3 & 50 & 264 & 109 & 16800 & 2400 & 4800 & 1825 & 262 & 522 \\
\bottomrule
\end{tabular}
}
\end{center}
\end{small}
\endgroup
\end{table*}

\section{ASSOCIATION FOOTBALL DATA}\label{sec:appendix_footbal_data}
A concise description of the marks $u$ used for our football case study in Section 4.1 is provided in Table \ref{table:football_description} while more details can be found in \cite{narayanan2021flexible}. 

\section{EXTRA EXPERIMENTAL RESULTS FOR THE FOOTBALL CASE-STUDY}

\begin{figure*}[ht]
\centering
\includegraphics[width=\linewidth]{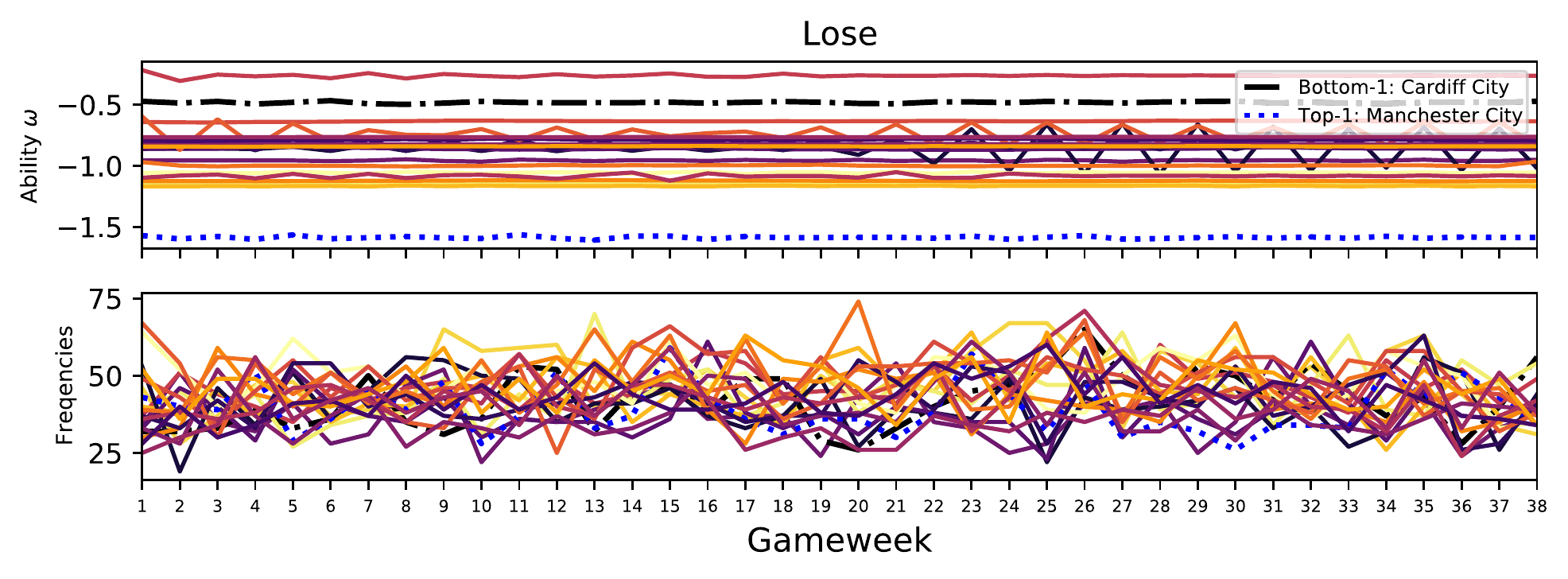} 
\caption{Top panel shows the evolution of losing the possession of the ball (Pass\_U) ability $\omega_{c, 9}^{(t)}$ for each team across the season. The blue dashed line corresponds to Manchester city, winner of 2013/2014 Premier league while the black dash-dotted line is for the bottom-ranked team Cardiff City. The colors for the rest of the teams have been assigned according to their final rank, with brighter colors associated with teams with a high rank at the end of the championship and darker ones to those having inferior ranks.}
\label{fig:omegas_lose}
\end{figure*}

We also illustrate similar results with Figure 2 of the main paper where now the event type ``Lose'' is taken into account. We see in top panel of Figure \ref{fig:omegas_lose} that almost all the teams have non-varying abilities through the season while their values are strongly related to final ranking of the teams. For instance, Manchester city, the winner of the championship, has the lowest ability values constantly across seasons. On the other hand, teams with the darkest colors which represent teams with low rank are found in the top positions. Once again as in the main paper, such patterns cannot be distinguished by the raw frequencies of the event per game week in the bottom panel.

\section{THE BRANCHING STRUCTURE}\label{sec:branching}
We use our proposed model of Section \ref{sec:autoregressive_model} to recover event genealogies using its hidden branching structure \cite{hawkes1974cluster}. The branching structure categorizes the events into immigrants and offsprings. Offspring events are triggered by previous events while immigrant events are not linked with a parent event. Let $w_i^{(s)}$ be the random variable indicating whether the $i$-th event of the $s$-th sequence is an immigrant ($w_i^{(s)} = 0$) or an offspring ($w_i^{(s)} = j$) of a previous event indexed by $j$, we can calculate analytically the conditional branching structure probabilities as 

\begin{align*}
 & p ( w_i^{(s)} = 0 | \cF_{t_i^{(s)}} ) = \frac{\delta_{u_i^{(s)}}^{z_i^{(s)}}}{\delta_{u_i^{(s)}}^{z_i^{(s)}} +\alpha \sum_{k: t_k^{(s)} < t_i^{(s)}} \gamma_{u_k^{(s)},u_i^{(s)}}^{z_i^{(s)}} \exp \left(-\beta_{u_k^{(s)},u_i^{(s)}}^{z_i^{(s)}} (t_i^{(s)} - t_k^{(s)}) \right)}, \\
 & p (w_i^{(s)} = j | \cF_{t_i^{(s)}}) = 
 \begin{cases}
       \frac{\gamma_{u_j^{(s)},u_i^{(s)}}^{z_i^{(s)}} \exp \left(-\beta_{u_j^{(s)},u_i^{(s)}}^{z_i^{(s)}} (t_i^{(s)} - t_j^{(s)}) \right) }{\delta_{u_i^{(s)}}^{z_i^{(s)}} +\alpha \sum_{k: t_k^{(s)} < t_i^{(s)}} \gamma_{u_k^{(s)},u_i^{(s)}}^{z_i^{(s)}} \exp \left(-\beta_{u_k^{(s)},u_i^{(s)}}^{z_i^{(s)}} (t_i^{(s)} - t_k^{(s)}) \right)}, &\quad\text{if}~~t_j^{(s)} < t_i^{(s)} \\
       0, &\quad\text{otherwise.} \\ 
\end{cases}
\end{align*}
The above probabilities are based on the  model in
\eqref{eq:mass_function_football} and their computation allows us to gain insights over the causality of the event occurrences by assuming a causal constraint that any event is triggered by exactly one of the previous events or the background. Hence, this calculation of probabilities attains the recovering of the hidden branching structure $w_i^{(s)}$.
We choose four different matches of the 2013/14 EPL winner Manchester City (MC) and we build the branching structure taking into account the last ten events before a MC's player scores the first goal in the first half of the game. More details are given in the caption of Figure \ref{fig:man_city_branching}. To gain better intuition of how this experiment is related to the real matches we provide the links of the videos with the goals scored from these four matches in the supplementary material. For the leftmost panel of Figure \ref{fig:man_city_branching}, we observe that our model suggests the events "A\_Clear`` and "H\_Dribble``, the first and third event, respectively,  before the first goal is scored, are the most probable event that leads to this goal. Interestingly, by watching the corresponding video, we observe that these two events play a crucial role for scoring this goal since MC's player Silva by dribbling/conveying the ball closer to the opponent's goal area creates the right circumstances for scoring the goal himself. The event "A\_Clear``, which is the most probable event that lead to a goal, is not unexpected since the main reason this goal was scored because a (failed) attempt of the opponent to move the ball away from his goal area. In the rightmost panel, the branching structure suggests that the goal was primarily a result of a player regaining possession of the ball from the opponent, i.e. the event "A\_Win``. The event "H\_Save`` also contributed to the triggering of the goal event. The video of the goal interestingly verifies that this goal was scored after  MC's player Jes\'{u}s Navas gained possession of the ball, leading to a fast counter-attack which was the main reason of the goal. The other event is also important since after the opponent's goalkeeper prevented a goal by the shoot of Edin Džeko, the ball landed at Silva's feet allowing him to easily score. It is encouraging that our model is able to capture such a level of detail in football dynamics while preserving interpretability.

\subsection{Links of videos}\label{sec:appendix_links} 
\begin{figure*}
\centering
\includegraphics[width=\linewidth]{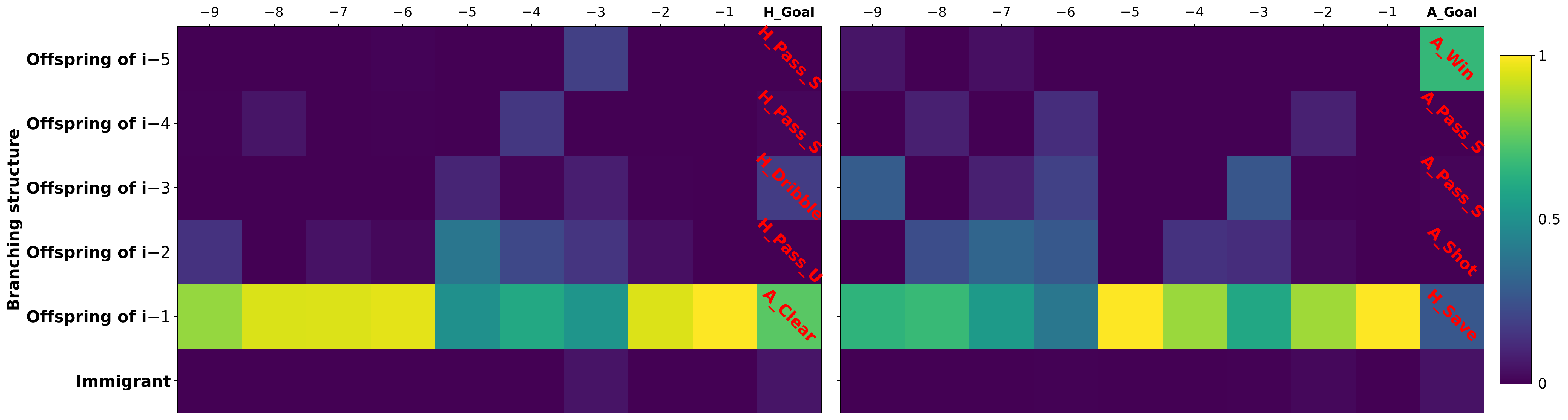}
\caption{Branching structure of the last 10 events before the first goal scored by Manchester City in the first half of the game from four different matches played by Manchester City (MC). The games are MC vs Newcastle United on 19/08/2013 (leftmost panel) and Arsenal vs MC on 29/03/2014 (rightmost panel). Each plot reads bottom-up with each column representing a probability vector which gives the probability the event on the x-axis is triggered by an immigrant or by an offspring that occurred at most 5 events in advance. The last column always depicts the sequence of the last 5 events (event types in red) before the MC scores.  The prefixes `H' or `A' indicates whether MC is the home or away team and the suffixes `U' and `S' indicate whether the passes were unsuccessful or successful, respectively.}
\label{fig:man_city_branching} 
\end{figure*}

The links of the videos for the two matches of Figure \ref{fig:man_city_branching} accompanied with the right time interval of the goal in the video in parentheses, the date of the match, the name of the MC's player who scored the first goal, and the final score.

\paragraph{MC vs Newcastle United. } \emph{Url}: \url{https://www.youtube.com/watch?v=ycnM_V273Zc&ab_channel=HDKoooralive} (see 0:00 - 1:00) ~~\emph{Date}: 19/08/2013 ~~\emph{Scorer's name of first goal}: Silva ~~\emph{Final score}: 4-0

\paragraph{Arsenal vs MC. } \emph{Url}: \url{https://www.youtube.com/watch?v=PQQqlGVb0Lk&ab_channel=mrszippy} (see 0:00 - 0:57) ~~\emph{Date}: 29/03/2014 ~~\emph{Scorer's name of first goal}: Silva ~~\emph{Final score}: 1-1

\vfill

\end{document}